# Large Hole Image Inpainting With Compress-Decompression Network

Zhenghang Wu    Yidong Cui

Beijing University Beijing University of Posts and Telecommunications

**Abstract**. Image inpainting technology can patch images with missing pixels. Existing methods propose convolutional neural networks to repair corrupted images. The networks focus on the valid pixels around the missing pixels, use the encoder-decoder structure to extract valuable information, and use the information to fix the vacancy. However, if the missing part is too large to provide useful information, the result will exist blur, color mixing, and object confusion. In order to patch the large hole image, we study the existing approaches and propose a new network, the compression-decompression network. The compression network takes responsibility for inpainting and generating a down-sample image. The decompression network takes responsibility for extending the down-sample image into the original resolution. We construct the compression network with the residual network and propose a similar texture selection algorithm to extend the image that is better than using the super-resolution network. We evaluate our model over Places2 and CelebA data set and use the similarity ratio as the metric. The result shows that our model has better performance when the inpainting task has many conflicts.

**Keywords**: Image Inpainting, Convolutional Neural Network, Compression-Decompression Network

## 1. Introduction

The image could be damaged by time, accidents, or irresistible force. Some parts of the image become stains. The image inpainting is applied to remove the invalid parts. After the breakthrough of the AlexNet [1] in the image classification field, a convolutional neural network is an essential approach in image processing.

The ruined image inpainting approaches that do not use convolutional networks also can repair the image. The approaches usually use mathematical methods to generate the result. For example, PatchMatch [2] method can patch images by reshuffling, which means using the image itself to fill the invalid holes. However, the approaches rarely use image semantics.

The convolutional neural network stacks many convolutional layers and uses activators and normalizations to generate suitable fitting weight parameters. These parameters can extract image semantics. Thus, using a deep convolutional network to inpainting is a reasonable choice. Pathak *et al.* [3] proposed the Context Encoders network that consists of two subnetworks. An encoder subnetwork compacts the image to feature matrixes, and a decoder subnetwork restores the missing part of the image.

Liu *et al.* [4] proposed Partial Convolution to modify the typical convolutional network, and the new network can well support irregular holes. Song *et al.* [5] proposed the Contextual-based Image Inpainting network in which they combine the convolutional network with color information. Yu *et al.* [6] proposed Gated Convolution to upgrade the mask updating method in Partial Convolution, and Gated Convolution also supports user-guided inpainting. Nazeri *et al.* [7]

proposed the EdgeConnect structure that uses image skeleton information before processing the inpainting operation.

The input of the image inpainting consists of two parts, an image and a mask directing which pixel is missing. Fig. 1 shows an example of the input. The top picture is the image matrix that has RGB channels, and the bottom picture is the mask matrix that has only one channel composed of 0 or 1. The value 0 (the black area) means missing pixels, and the value 1 (the white area) means valid pixels.

The existing approaches can patch the image with partial missing (as shown in Fig. 1), but when the missing part becomes larger, the output will exist blur and color confusion. We study the existing approaches and propose a new deep convolutional neural network, Compression-Decompression network. In a way, the network can patch the edge-hole image.

The existing models mainly use an encoder-decoder structure that has two subnetworks, an encoder network, and a decoder network. The encoder network takes responsibility for extracting feature information from input images, and the decoder network is in charge of reconstructing the image from the information. However, the encoder-decoder structure has bad performance in edge-hole image inpainting. Our model has two subnetworks referring to the encoder-decoder network, a compression network, and a decompression network. The difference between the encoder-decoder and compression-decompression is that the output of the compression is the down-sample image similar to the ground truth.

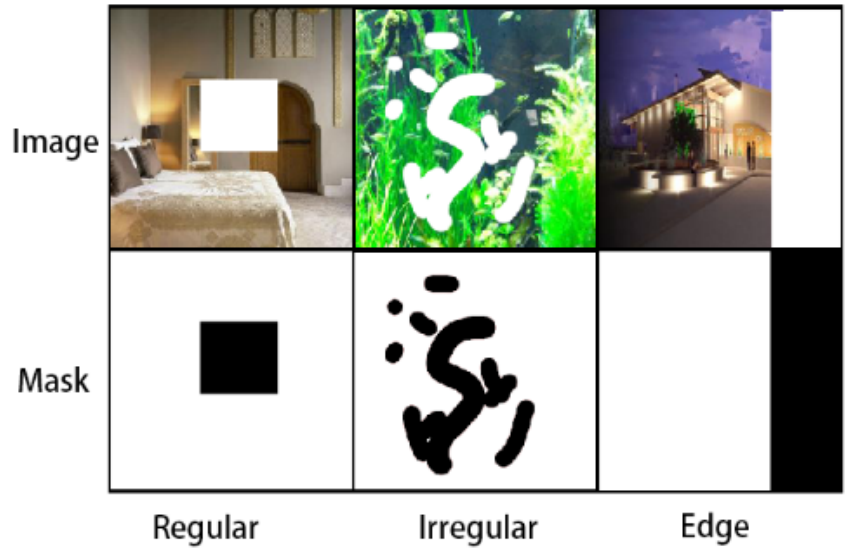

Fig. 1. An example of the image inpainting. The regular mask means the rectangular hole. The irregular mask means the random hole. The edge mask means around 30% of pixels are missing at the edge of the image.

Our main contributions can be summarized as follows:
- We propose a new network structure to patch the edge-hole missing. Compared with the encoder-decoder network, the compression-decompression network consists of two subnetworks. The network has better performance in Places2 [18] data set.
- The compression-decompression network consists of two subnetworks. The compression network is a ResNet-like network that uses the gated convolution layer [6]. The decompression network is a super-resolution network that enlarges the output of the compression network.
- We propose a new approach to enlarge the output, a similar texture selection algorithm. The

algorithm selects the most similar pixel between the output and the input image. It improves the performances over the super-resolution network.

## 2. Related Works

Pathak *et al.* [3] proposed the encoder-decoder convolutional neural network, Context Encoders, to patch the regular hole in the image. The encoder-decoder network consisting of two subnetworks. The first network, the encoder network, is used to compress the image matrix and export feature matrixes that contain high-level image semantic information. Moreover, the second subnetwork takes responsibility for extending the output feature matrixes.

Liu *et al.* [4] proposed the partial convolutional layer that processes the image matrix and the mask matrix separately compared to the typical convolutional layer. The forward step of the mask matrix is using a sliding window. If the sum value of the values in the window is larger than zero, the output value is one. That is $WindowOutput = Min(Sum(SlideWindow), 1)$. The output then is multiplied by the convolutional output of the image matrix, which means only the edge of the invalid hole is repaired in every layer. That is $Output = Conv(Image) \odot WindowOutput$.

Yu *et al.* [6] proposed the gated convolutional layer similar to the partial convolutional layer. Compared to the partial convolution, the gated convolution uses a typical convolution to process the mask matrix instead of using the sliding window. So, the network can update the weight parameters by the back-forward algorithm and allow the user-guided mask that means users can predefine some inpainting information before inpainting. The output becomes $Output = Conv(Image) \odot Conv(Mask)$.

Song *et al.* [5] proposed the contextual-based inpainting network. The network divides the inpainting into three tasks: inference task, match task, and translation task. The inference task uses an Image2Feature network to extract feature information. The matching task uses swapping operation to find similar features in the valid area of the image. The translation task restores the feature to the image.

Nazeri *et al.* [7] proposed the edge-connect inpainting network. In the network, Nazeri *et al.* construct a new subnetwork to patch the edge information (stored in edge mask matrixes) of the damaged image before starting inpainting. The network uses the Canny edge detector algorithm [8] to extract the edge of the true image.

The existing approaches use the GAN [9] network as the main loss generator and use the L1 loss, the L2 loss, and the feature loss of the VGG net [10] as the additional loss to accelerate the training of the GAN network. The VGG feature loss means use the pre-trained VGG network, choose the output of some high-level layers, and apply the L1 or L2 loss to the output between the generated image and the ground truth.

## 3. Compression-Decompression Network

We propose a convolutional neural network to inpainting the edge hole missing image, the compression-decompression network that consists of two stages: the compression stage and the decompression. In the network training, we use a GAN network to discriminate whether the down-sample output is close to the true image and use the variance loss to smooth the output color.

## 3.1. Compression Subnetwork

The compression subnetwork is the same as ResNet[11] network. The difference is that the compression network uses the gated convolution layer [6] that has a new convolution layer to process mask matrixes. According to the structure of ResNet, the network consists of a stack of the gated layers and the residual layers. The compression network comprises three residual blocks that are shown in Table 1. RB (Residual Block) represents the residual operation layer that recasts the normal convolution mapping. CB (Convolution Block) represents a sequential stack of gated convolution layers. MB (Mapping Block) represents a mapping layer that transforms the feature matrix into the output image. One RB and one CB are a pair that the output of RB will be added into the output of CB, which is similar to ResNet. LeakyReLU [12] and BatchNormal [13] are used in all gated layers. For example, the output of RB_0 will be elementwise-add to the output of Conv_4 in the pair of RB_0 and CB_0.

| Block_Name | Layer_Name | Kernel_Size | Channel_Number | Stride |
|---|---|---|---|---|
| RB_0 | - | 1 | 64 | 2 |
| CB_0 | Conv_0 | 5 | 64 | 2 |
|  | Conv_1 | 5 | 64 | 1 |
|  | Conv_2 |  |  |  |
|  | Conv_3 |  |  |  |
|  | Conv_4 |  |  |  |
| RB_1 | - | 1 | 196 | 2 |
| CB_1 | Conv_5 | 5 | 196 | 2 |
|  | Conv_6 | 5 | 196 | 1 |
|  | Conv_7 |  |  |  |
|  | Conv_8 |  |  |  |
|  | Conv_9 |  |  |  |
| RB_2 | - | 3 | 256 | 2 |
| CB_2 | Conv_10 | 3 | 256 | 2 |
|  | Conv_11 | 3 | 256 | 1 |
|  | Conv_12 |  |  |  |
|  | Conv_13 |  |  |  |
|  | Conv_14 |  |  |  |
|  | Conv_15 |  |  |  |
|  | Conv_16 |  |  |  |
| MB | - | 3 | 3 | 1 |

Table 1. The structure of the compression network.

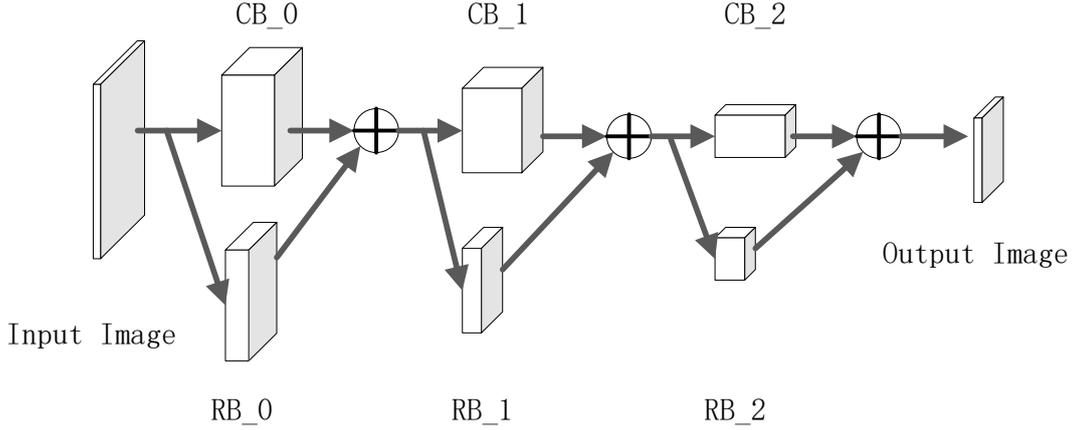

Fig. 2. The data flow of the compression network. The "⊕" represents element-wise adding.

In the training step, we use Adam Optimizer [14] with learning rate=2e-4 and Spectral-Normalized Markovian Discriminator (SN-PatchGAN) [15] that provides GAN-Loss to discriminate whether the output is similar to the ground truth. The additional L1 loss is used to guide the learning optimizer.

## 3.2. Decompression Subnetwork

After the processing of the compression network, the invalid image becomes a down-sample image. In our network, the input image size is 256x256, and the down-sample image size is 32x32. The main responsibility of the decompression network is extending the down-sample image. We use two approaches to undertake the task.

**Super Resolution**. The super-resolution technology uses a deep convolutional neural network to extend the image resolution. We select EDSR [16] as the network. In the paper [16], the max extension scale is x4, but the 32x32 size image needs the scale x8. So, we stack two EDSR as an x8-scale super-resolution network. The output image still losses many texture details, because the scale is too large.

**Similar Texture Selection**. Although the super-resolution network can enlarge the down-sample image, the extended image still losses many texture information, because it is hard to extend the image from 32x32 to 256x256 resolution.

We propose a new approach to alter the scale, a similar texture selection approach. The similar texture selection uses the down-sample output and the true resized image. The algorithm compares every pixel between images, finds the most similar pixel in the true low-resolution image, and selects the corresponding texture block in the true high-resolution image.

Formula (1) shows how to find the most similar pixel. Fig. 3 shows an example. LROutput represents the low-resolution output of the compression network. LRGT represents the true low-resolution image. The subscripts, x, y, i, and j, locate the pixels in the image. The loss function is L1 loss in RGB channels. After finding i and j, the algorithm starts the copying stage. HROutput represents the high-resolution output of the decompression network. Formula (2) shows the copying operation. The subscripts, n, and m, represent the copying index whose range is from 0 to 8. The range directs the block size that is calculated by 256/32=8.

After the selection stage, the algorithm finetunes the output based on the input image to add more texture information. We firstly stretch the damaged image and calculate the loss between the output and the stretched part. If the loss is less than a threshold value (a pre-defined hyperparameter), we will replace the pixel of the output with the pixel of the stretched image. Formula (3) shows how to calculate the output. The parameter t represents the threshold value. The subscripts, i and j, represent the index of one pixel. Fig. 4 shows an example.

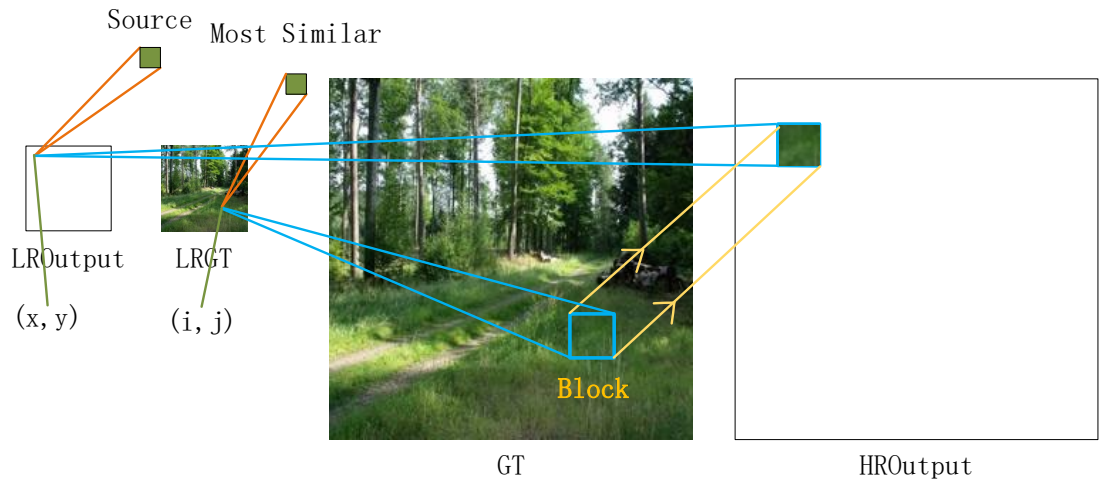

Fig. 3. An example of the similar selection algorithm. The orange line locates the pixel. The Index is (x, y) in the LROutput, and the most similar index is (i, j) in the LRGT. The blue line locates the corresponding pixel block. The yellow line represents copying operation. The green line shows the index.

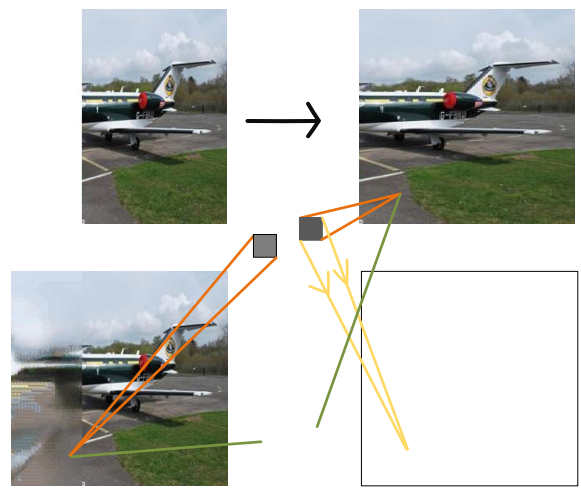

Fig. 4. An example of how to finetune from the stretched image. The orange line locates the pixel whose index is (i, j). The yellow line represents copying the pixel if the loss is less than the threshold value.

$$i,j = \mathop{Argmin}_{i,j} \left\{ \left| LROutput_{x,y} - LRGT_{i,j} \right|_{loss}, 0 \leq i,j < 32 \right\} \quad (1)$$

$$HROutput_{x \times 8+n, y \times 8+m} = GT_{i \times 8+n, j \times 8+m} \quad 0 \leq n,m < 8 \quad (2)$$

$$Output = \begin{cases} Stretched_{i,j}, & if\ |HROutput_{i,j} - Stretched_{i,j}| < t \\ HROutput_{i,j}, & otherwise \end{cases} \quad (3)$$

## 4. Experiments

### 4.1. Comparisons

We use Pytorch [17] as the calculation platform and Places2 [18] image data set as the input image (256x256). Four missing masks are used to represent the top, bottom, left and right missing. The masks are selected randomly in the training stage.

We compare our method with three methods:

-GC [5]: The network uses the gated convolutional layers.

-PC [4]: The network uses the partial convolutional layers

-EC [7]: The network has an additional network. The edge inpainting network patches the edge of the missing part before starting inpainting

-Our: Our compression-decompression network that uses the similar texture selection algorithm.

Figure 5 shows the result.

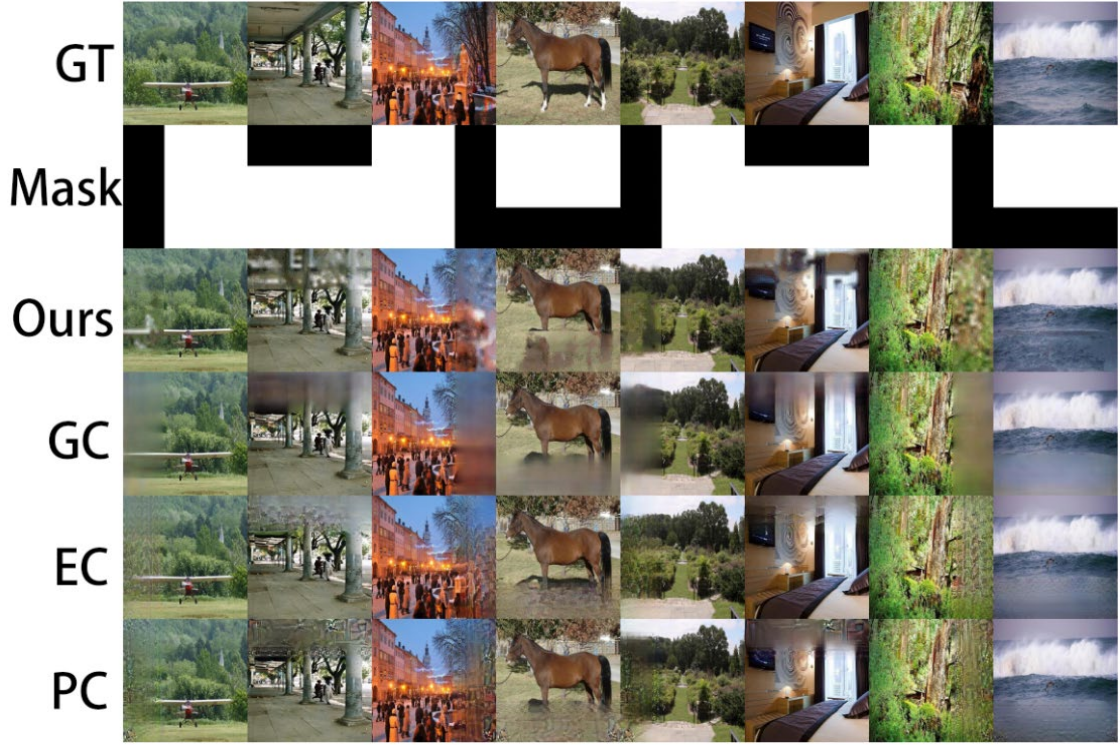

Fig. 5. Comparisons of testing results on Places2 data set.

We use the L1 loss as the output metric, but compared to the original inpainting task, the large hole missing inpainting task is more like a generative task. The main purpose is that the output image should be more similar to a real image instead of more similar to the input ground truth. So, we use a new metric, the similarity ratio, to measure whether the output is acceptable.

We use a classification convolutional network to calculate the similarity ratio. Formula (4) shows how to calculate the ratio. CN represents the convolutional network, and CN(.) represents the output of CN. IndexMax(.) represents selecting the index of the largest value in the input. I(x, y) function represents a discriminator that if the input x is equal to the input y, it returns 1, otherwise returns 0.

$$Similarity = \frac{1}{n}\sum_{i=0}^{i<n} I\Big(IndexMax\big(CN(HROutput_i)\big), IndexMax\big(CN(GT_i)\big)\Big) \quad (4)$$

In the experiment, we use a pre-trained VGG16_BN network as the discriminator and add a new full-connection layer with 365 channels to transfer from 1000-classification task to 365-classification task. After finetuning, we can calculate the similarity ratio. Table 2 shows the result.

|  | L1 Loss | L2 Loss | Similarity | Similartiy5 |
|---|---|---|---|---|
| Ours_STS | 0.0609 | 0.0213 | **0.64011** | **0.91024** |
| Ours_SRN | 0.0611 | 0.0215 | **0.62934** | **0.90646** |
| GatedConv | 0.0513 | 0.0148 | 0.62904 | 0.89172 |
| PartialConv | 0.0520 | 0.0156 | 0.54556 | 0.86786 |
| EdgeConnect | **0.0321** | **0.0146** | 0.32021 | 0.65375 |

Table 2. Comparisons with various methods. The similarity5 represents if the index is in the top-5 index of the VGG output of the ground truth. The Ours_STS represents using the similar texture selection algorithm. The ours_SRN represents using the super-resolution network.

## 4.2. Discussion

So why the similarity ratio of the compression-decompression network is higher than the other encoder-decoder-like network? We think it is because of the conflict between the output and the ground truth. Since the output resolution of the encoder-decoder is the same as the input image resolution, and the output of the compression-decompression is low-resolution, the down-sample output alleviates the conflict when we calculate the loss because the larger output image means the larger coverage range of the convolution kernel. A convolution kernel processes a block by weight sharing. Thus, the 32x32 resolution only needs to fit 1.5625% (32*32/256/256) pixels when the optimizer trains the kernel weights. For example, the pixel A needs RGB=(0.0, 0.0, 0.0) and the pixel B needs RGB=(1.0, 1.0, 1.0). But the input of the pixels is almost equal. Thus, the convolution kernel cannot fit the input. Because one input only corresponds to one output.

The second question: why does the conflict disappear in the original inpainting task? Unlike the original inpainting task, the new task demands the inpainting network to reconstruct objects, including edges, color, textures, and shadow in the large hole. The hole is larger, the conflict is more. Moreover, the original task can use the surrounding pixels since the hole is not too large. On the contrary, the new task can only use the adjacent pixels. This condition gives the network more useful information than the new task.

Thus, the compression-decompression network has better performance. However, if the inpainting task has less conflict, the original network becomes better, because in the decompressing stage, the compression-decompression network losses many information (the output of the compression network has only three channels).

We use CelebA[19] data set as the less conflict testing data set since all images are human face images (Places2 has 356 classes). We can see that the output of the existing approaches becomes better. On the contrary, the compression-decompression network needs to enlarge the down-sample image. This operation causes detail losing, and the CelebA is a one-class data set that has little conflict when we calculate the loss.

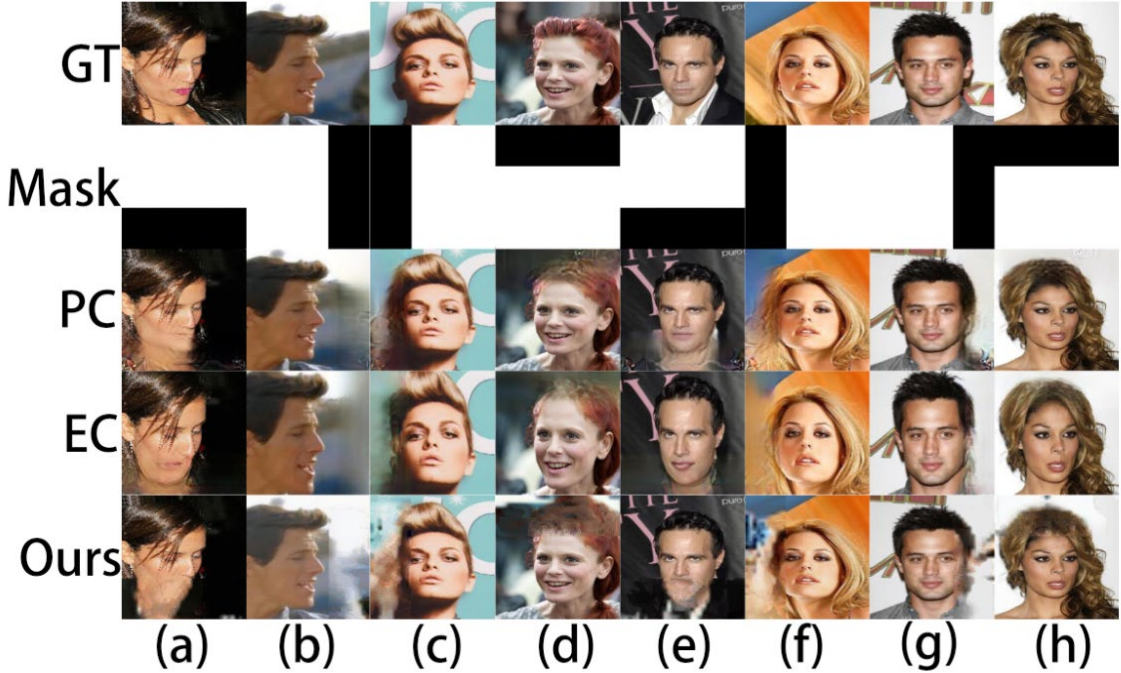

Fig. 6. Comparisons of testing results on CelebA data set. The (e) image shows that the output of EdgeConnect is better than the compression-decompression network.

## 5. Future Works

According to the output of CelebA data set, we can see the compression-decompression network still has some limitations: if the image in the data set is highly similar, the output of the network is worse than the other networks because of the information loss in the compression stage. Thus, our future work is trying to construct a network to restore the information.

## 6. Conclusion

In the paper, we aim to extend the range of the inpainting task and reconstruct the network to make it possible to patch the large vacant image. The existing approaches use the encoder-decoder structure, but when the input is a missing edge image, the output image has color blur and confusion. Thus, we construct the compression-decompression network consisting of two subnetworks. The compression subnetwork takes responsibility for the inpainting task, and its output is down-sample images. The decompression network is in charge of resizing the output image into the original resolution. In the experiment, we use Places2 and CelebA data set and propose the similarity ratio metric that can measure whether the output is similar to the ground truth. The compression-decompression network has better performance in Places2, but worse in CelebA. In the end, we analyze the reason why it happens: it is because Places2 images have more conflict when we calculate the loss between output images and true images, compared to CelebA.